\documentclass{article}
\usepackage{spconf,amsmath,graphicx,hyperref}

\usepackage{url}
\usepackage{xcolor}
\usepackage{booktabs}
\usepackage{multirow}
\usepackage{amssymb}

\graphicspath{{figures/}}

\title{WaveCommitments: Private and Verifiable Speech-Agent Marketplaces}

\name{Oguzhan Baser$^{\dagger *}$ \qquad Kaan Kale$^{\ddagger}$ \qquad
      Sriram Vishwanath$^{\ddagger}$ \qquad Sandeep Chinchali$^{\dagger}$
\thanks{$^{*}$Corresponding author. Email: oguzhanbaser@utexas.edu.\\
This work was supported in part by the National Science Foundation under Grant 2148186.
Any opinions, findings, and conclusions or recommendations expressed in this material are those of
the authors and do not necessarily reflect the views of the National Science Foundation.}}

\address{$^{\dagger}$The University of Texas at Austin, Austin, TX, USA \\
         $^{\ddagger}$Georgia Institute of Technology, Atlanta, GA, USA}

\begin{document}
\ninept
\maketitle

\begin{abstract}
Modern speech systems are evolving into ecosystems of autonomous speech agents that interact, negotiate, and trade assets such as embeddings, spectrograms, or raw data. In such markets, sellers cannot reveal assets for evaluation due to privacy constraints, and buyers cannot price what they cannot test. Thus, we introduce WaveCommitments, an end-to-end design that couples multivariate polynomial commitments with confidential asset valuation via Data Shapley inside a trusted execution environment. It enables selective openings and tamper-evident provenance to prove that a delivered asset matches the valuation while hiding everything else. Our extensive experiments across three standard datasets and various tasks with state-of-the-art encoders, including Whisper, reveal that we speed up the strongest baselines by up to 18.1$\times$ while reducing their proof sizes by up to 43.2$\times$. We also analyze how a speaker's value shifts across tasks and encoders, revealing agent-level market prices.
\end{abstract}


\section{Introduction}

Modern speech systems shift from standalone models to ecosystems. Imagine that speaker encoders, clinical audio diagnostic tools, and proprietary vocal models are owned, updated, and deployed by different parties. These parties, often called \textbf{autonomous speech agents}, interact and negotiate resources on the fly without humans in the loop. In this regime, marketplaces for speech knowledge - utterances, spectrograms, embeddings, and adapters - become inevitable. Speech foundation models benefit from data at the scale of tens to hundreds of thousands of hours~\cite{whisper,wav2vec2citation}, yet \textit{the data that benefits them most in practice is often the hardest to share} (e.g., consented medical audio, copyrighted vocals, and accent-specific collections under privacy constraints). Even when raw audio cannot move, compute becomes the bottleneck: extracting high-quality representations from large speech-language models requires substantial GPU effort. Consequently, deployments need mechanisms that let agents \textit{buy, sell, and reuse} proprietary or private speech knowledge reliably with incentives for honest data curation.

\begin{figure}[t]
    \centering
    \includegraphics[width=\linewidth]{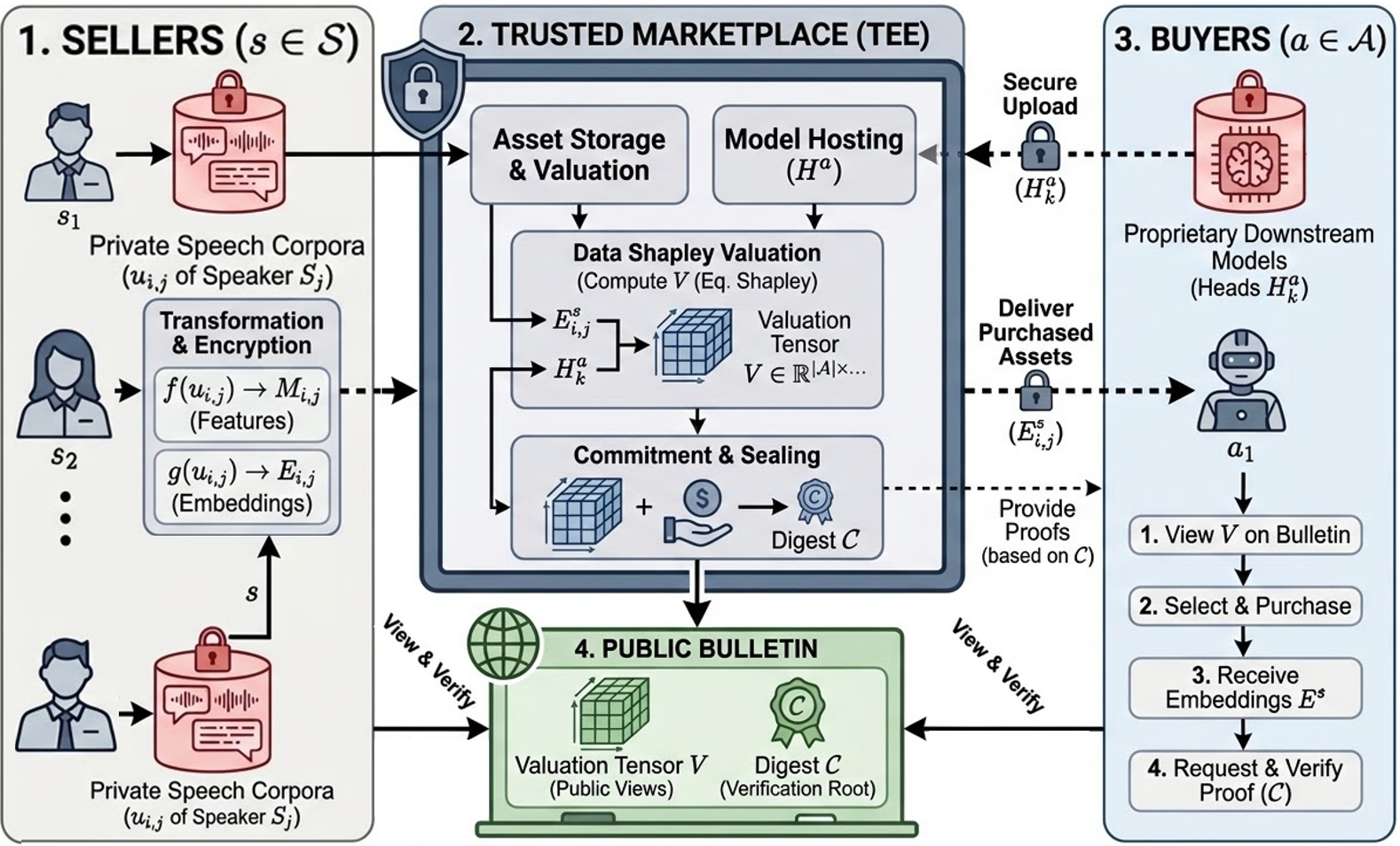}
    \caption{\textbf{WaveCommitment Architecture:} Imagine a SAM where autonomous agents trade proprietary speech knowledge \textit{(red)}, such as embeddings. Seller $s$ cannot reveal raw utterance $u$, and buyer $a$ cannot trust unseen embeddings $E^s$ or determine their value $V$ for its model $H^a$. Hence, they both upload their assets to a secure enclave which computes the valuations for each pair confidentially, seals those and the referenced assets into a single digest $C$ to publish in a public bulletin \textit{(green)}. After $a$ and $s$ handshake based on $V$, $a$ buys $E^s$ and uses $C$ to validate it without exposing other assets.}
    \label{fig:sysarchitecture}
\end{figure}

Despite this need and the potential of speech agent marketplaces (SAMs), a \textit{central open problem} is \textbf{trust under confidentiality}: sellers must convince buyers that speech assets are authentic and relevant without exposing the very information that makes them valuable, while buyers must price assets they cannot freely test in their own pipeline. Existing work falls into three categories, each insufficient on its own. First, \textit{reveal-to-evaluate workflows} share utterances or embeddings for trial~\cite{revealtoevaluate, distributedleakage}. They undermine privacy and consent for sensitive audio (e.g., clinical acoustics, children's speech, or proprietary calls) and for copyrighted content (e.g., audiobooks, vocals)~\cite{distributedleakage, xvecdeanonymization, embeddingleakage}. Second, \textit{metadata-and-reputation signals} (e.g., diversely curated accent, verified speakers) are too weak for speech, as utility is highly model- and domain-specific~\cite{datahshapley, augmenteddatasheets}. For example, an accent dataset that boosts one speaker recognition (SR) model may be marginal for another; a speaker embedding bank may help diarization but not robustness. Moreover, coarse summaries cannot capture whether samples actually benefit the buyer's objective. Third, \textit{cryptographic mechanisms}~\cite{merkle, chen2024zkml, zkllm} offer stronger guarantees, but often at high overhead or slow verification, which is challenging when SAMs operate at the granularity of tens of thousands of utterances or embeddings. In addition, SAMs face uniquely acute integrity risks: synthetic speech and embedding forgeries can look ``plausible'' while being low-diversity or strategically poisoned~\cite{baser2025phonemefake}, and even ``benign'' embeddings can leak identity or content~\cite{baser2025wavshape}. What is missing for SAMs is a practical mechanism that jointly provides \textit{(i)} confidential, model-specific valuation, \textit{(ii)} provenance against manipulation, and \textit{(iii)} scalable verification.

We \textbf{observe} that speech ``knowledge'' assets are inherently \textit{multi-dimensional tensors} (e.g., spectrograms, filter banks, embedding matrices). When current solutions treat them as uni-dimensional, they create a mismatch: verifying many individual entries is too slow, while blindly analyzing individual segments leaks correlated information (e.g., phonetic, prosodic, or speaker cues). Our \textbf{key insight} is that verifiability must be \emph{multivariate by design}. Thus, we propose WaveCommitments (WC): a \emph{fingerprint} of a spectrogram or embedding tensor that is published once, then later allows anyone to \textit{open} and check only part of a spectrogram or utterance, without revealing the rest. WC commits to speech assets in their native tensor form and supports \emph{selective openings} that prove authenticity and correct speech/embedding derivation only for the queried entries, while keeping all other regions hidden. WC is based on multivariate polynomials and integrates with a marketplace backed by a trusted execution environment (TEE) performing a confidential and model-specific evaluation as in Fig.~\ref{fig:sysarchitecture}. In short, WC is an end-to-end system for confidential valuation and verification of speech assets under privacy constraints.

\textbf{\underline{Related work}} addresses the trust problem only partially. \textit{TEE or MPC based private speech inference}~\cite{voiceguard2018,teixeira2022privateasr,chouchane2021antispoofingmpc} protects audio during compute but produces no checkable record tying a delivered embedding to its agreed valuation. \textit{Watermarking}~\cite{baser2024securespectra, icmlwatermark} attributes content but cannot verify that embeddings are from real speech. \textit{Cryptographic methods} involving Merkle/Verkle trees~\cite{merkle,verkle,kzg} and zero-knowledge proofs~\cite{chen2024zkml, zkllm} offer integrity guarantees but treat tensors as flat lists, triggering large size or delayed proofs. WC bridges all three by performing confidential model evaluation inside a TEE, and publishing \textbf{low-cost verification} objects via \textbf{multivariate commitments} that preserve the speech geometry. In light of prior work, our \textbf{contributions are fourfold:}
\begin{itemize}
    \item We build a \textbf{confidential speech-agent marketplace} where \textit{agents trade speech knowledge} (e.g., features or embeddings) and the market \textit{privately} computes model-aware prices via \textit{Data Shapley} inside a \textit{TEE}.
    \item We formulate \textbf{a multivariate commitment} tailored to \textit{speech agents}. A \textit{public digest} ``seals'' speech and valuations to verify the match later with \textit{a small proof}.
    \item We build a \textbf{hierarchical tree} on our commitments. It \textit{organizes} each speaker's utterance, spectrogram, and embeddings, so agents can verify a chosen neighboring subset (e.g., a speaker's high-pitched band) scalably.
    \item We release our code\footnote{\href{https://github.com/UTAustin-SwarmLab/WaveCommitments}{\texttt{github.com/UTAustin-SwarmLab/WaveCommitments}}\label{code}} to catalyze trustworthy speech markets.
\end{itemize}

\section{Methodology}

Imagine a \emph{SAM} where speech agents under confidentiality constraints trade \emph{speech knowledge} involving utterances, spectrograms, embeddings, and LoRA-like adapters. \textbf{Four entities} exist in the marketplace. \textit{Seller} $s$ in a set of sellers $\mathcal{S}$ owns private speech corpora or derived representations. \textit{Buyer} $a$ among buyers $\mathcal{A}$ owns proprietary models and seeks data to improve task performance (e.g., SR, clinical acoustic prediction). \textit{Trusted Marketplace} runs inside a \textit{TEE} and executes a valuation pipeline over private seller and buyer assets. \textit{Public Bulletin} stores the valuations and a small verification digest.\\

\begin{figure}[t]
    \centering
    \includegraphics[width=\linewidth]{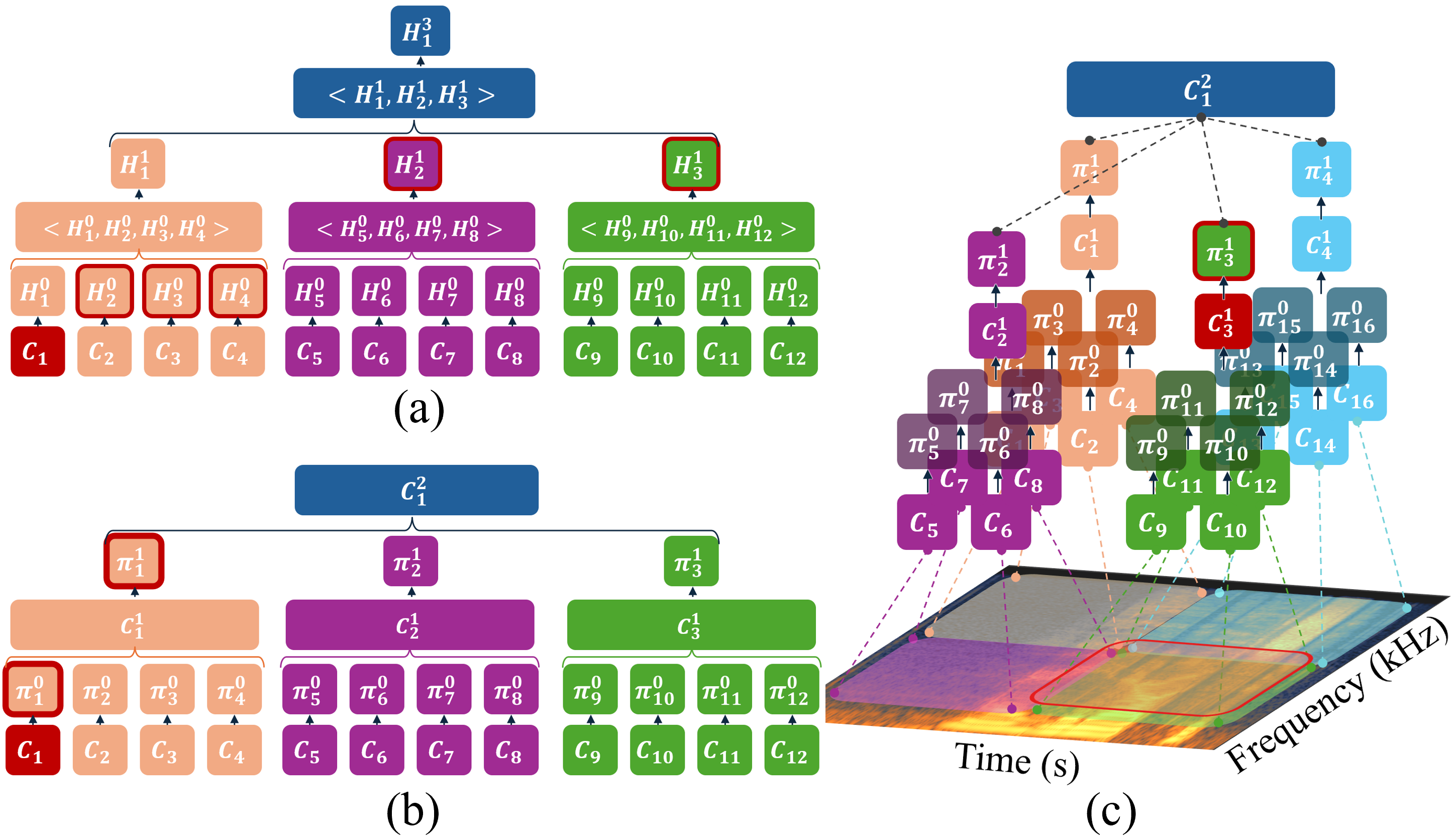}
\caption{\textbf{How to selectively verify speech tensors?} WaveTrees (c) organize assets in native time-frequency geometry and return a short proof for the region of interest, unlike Merkle (a) and Verkle (b) trees with larger overhead (outlined boxes). Ours can verify whether a speaker says a phrase in a time window while copyrighted music on background frequencies remains hidden.}
    \label{fig:WaveTree}
\end{figure}

\noindent\textbf{\underline{Problem Formulation:}} Each seller $s$ has $n_j$ private utterances $u_{i,j}$ of speaker $S_j$, one of the $\gamma_s$ speakers held by that seller, where $i\in[n_j]$ and $j\in[\gamma_s]$. The seller transforms the asset $u_{i,j}$ as: $M_{i,j} = f(u_{i,j}),E_{i,j}=g(u_{i,j})$, where $M_{i,j}$ denotes features (e.g., MFCC/filterbank) and $E_{i,j}\!\in\!\mathbb{R}^d$ denotes an embedding or spectrogram tensor of dimension $d$. Then, $s$ uploads $E_{i,j}^s$ (or its derivatives) to the TEE. Each buyer $a$ has several prediction or mixture-of-expert heads $H^a_k(\cdot)$ among $h_a$ heads, $k\in[h_a]$, and uploads those to the TEE such that the model parameters remain hidden outside the enclave. Next, the TEE computes a valuation tensor $V\!\in\!\mathbb{R}^{ (\sum_{a\in\mathcal{A}} h_a)\times (\sum_{s\in \mathcal{S}}\sum_j^{\gamma_s}n_j)}$ by Data Shapley given in Eq.~\ref{eq:shapley_def} and tabulates $V$ indexed by asset identifiers \((s,j,i,a,k)\). Finally, it seals $V$ and the assets by committing to a compact digest $C$ and releases $C$ and $V$. The buyer $a$ then selects the entries to purchase based on the values in $V$ and receives the corresponding representations (typically $E_{i,j}^s$, not $u_{i,j}^s$). Later, buyer $a$ requests short proofs that the delivered $E_{i,j}^s$ is exactly the committed representation
without seeing any other assets in non-purchased regions.\\

\noindent\textbf{\underline{Threat Model:}} SAMs without WCs face sharp integrity and privacy attacks. Sellers can \textit{substitute} a post-bid embedding or \textit{forge} plausible synthetic assets lacking provenance or diversity. Buyers can \textit{free-ride} via evaluation access, \textit{adaptively} \textit{poison} seen assets by reversing their own model to harm competitors, or \textit{leak} speaker cues from repeated disclosures.\\

\noindent\textbf{\underline{Assumptions:}} We assume the TEE correctly runs the agreed code and enforces confidentiality against the outside host at the software boundary. Our cryptographic layer provides \emph{non-equivocation} and \emph{tamper-evidence} once a public digest is posted (i.e., a committed entry cannot be reopened to another value), matching the standard binding in commitments. Side-channel and physical compromise defenses are orthogonal, treated as platform-level mitigations.\\

\noindent\textbf{\underline{Data Shapley (DS):}} For each buyer agent $a\in\mathcal{A}$, the market must assign a \emph{model-specific} value to each candidate asset (e.g., embedding $E$ or a tensor) without revealing the buyer weights or the seller's raw speech. We use DS as the valuation primitive. Let $\mathcal{D}=\{z_1,\dots,z_N\}$ denote the set of $N$ tradable data items (e.g., $z_l=E_{i,j}$ or $(M_{i,j},E_{i,j})$). For buyer agent $a$, let $\psi(a,\mathcal{Z})$ be a lightweight fine-tuning protocol producing  model $\theta_{a,\mathcal{Z}}^*$ from subset $\mathcal{Z}\subseteq\mathcal{D}$. Let $\varepsilon_a(\theta_{a,\mathcal{Z}}^*)$ be the evaluation metric of agent $a$ on its private held-out set (e.g., task accuracy), so that higher values are better. We define agent $a$'s buyer utility $U_a$ for a subset $\mathcal{Z}$ as $U_a(\mathcal{Z}) :=\varepsilon_a(\theta_{a,\mathcal{Z}}^*)$.

\textit{Definition 1: The Shapley value $\phi_{a,l}$ for data $z_l$ (with respect to an agent $a$) is the expected marginal utility $U$ of adding $z_l$ to a uniformly random ordering of all items, given as:}
\begin{equation}
\phi_{a,l}:=\frac{1}{|\zeta|}\sum_{p \in \zeta}U_a(P^p_l \cup \{z_l\}) - U_a(P^p_l),
\label{eq:shapley_def}
\end{equation}
\textit{where $P^p_l:=\left\{z_{p(r)}:r<p^{-1}(l)\right\}$ is the set of items preceding $z_l$ in permutation $p$ and $\zeta$ is the set of all permutations of $\mathcal{Z}$}.\\
In WC, the market publishes the valuation tensor $V$ with entries $V_{a,l} := \phi_{a,l}$ together with a root binding all $(a,l,V_{a,l})$.\\

\noindent\textbf{\underline{Monte-Carlo Estimation of DS:}}
Exact calculation of Eq.~\ref{eq:shapley_def} is $O(|\zeta|)$. We estimate $\phi_{a,l}$ by sampling $K$ random permutations $p^{(1)},\dots,p^{(K)}$ and averaging marginal contributions, as in:
\begin{equation}
    \hat{\phi}_{a,l}:= \frac{1}{K}\sum_{k=1}^K U_a\big(P^{p^{(k)}}_l \cup \{z_l\}\big) - U_a\big(P^{p^{(k)}}_l\big).
\label{eq:shapley_mc}
\end{equation}

For each permutation in Eq.~\ref{eq:shapley_mc}, the TEE builds the prefix set incrementally and evaluates $U_a(\cdot)$ along the prefix. Practically, the TEE can \emph{warm-start} training between successive prefixes and apply \emph{truncation} when the incremental gain becomes negligible for long stretches, reducing the number of expensive $\psi_a,\varepsilon_a$ calls while preserving an accurate estimate of DS.\\

\noindent\textbf{\underline{WaveCommitments:}} We work over a finite field $\mathbb{F}$ and pairing-friendly groups $\mathcal{G}, \mathcal{G}_T$ of prime order equipped with a bilinear map $e:\mathcal{G}\times\mathcal{G}\to\mathcal{G}_T$ and generator $G\in \mathcal{G}$~\cite{bls}. A WC binds an order-$m$ tensor $T\in \mathbb{F}^{d_1\times\cdots\times d_m}$ to a single group element $C_T\in \mathcal{G}$ while enabling succinct openings of individual tensor entries indexed by $\omega=(\omega_1,\dots,\omega_m)$. Given tensor shape $d=(d_1,\dots,d_m)$, WC samples independent trapdoors $\tau_1,\dots,\tau_m\in\mathbb{F}$ and publishes a structured reference string (SRS) containing group elements for all terms up to degree $d_j-1$ along each axis $j$. For a multi-degree $\textbf{i}=(i_1,\dots,i_m)$ with $0\le i_j\le d_j-1$, define $\tau^\textbf{i} := \tau_1^{i_1}\cdots \tau_m^{i_m}$ and $G_\textbf{i}:=G^{\tau^\textbf{i}}$. The public SRS is $\{G_\textbf{i}\}_{\textbf{i}\in I}$ where $I:=\{0,\dots,d_1-1\}\times\cdots\times\{0,\dots,d_m-1\}$.
Given tensor values $T[\omega]$ on grid $\Omega$, the TEE interpolates a multivariate polynomial $F_T\in \mathbb{F}[X_1,\dots,X_m]$ with per-variable degree bound $d_j-1$ so that $F_T(\omega)=T[\omega]$ for all $\omega\in\Omega$, where $T[\omega]=z$. It writes $F_T$ in the monomial basis, as in:
\begin{equation*}
F_T(X_1,\dots,X_m)
=
\sum_{k_1=0}^{d_1-1}\cdots\sum_{k_m=0}^{d_m-1}
c_{k_1,\dots,k_m} X_1^{k_1}\cdots X_m^{k_m},
\end{equation*}

and commits by evaluating $C_T
:= G^{F_T(\tau_1,\dots,\tau_m)}$, shown as:
\begin{equation*}
G^{F_T(\tau_1,\dots,\tau_m)}
=
\prod_{k_1=0}^{d_1-1}\cdots\prod_{k_m=0}^{d_m-1}
\left(G^{\tau_1^{k_1}\cdots \tau_m^{k_m}}\right)^{c_{k_1,\dots,k_m}}
\in \mathcal{G}.
\end{equation*}

We compute it using the public SRS via linearity in the exponent. To open entry $\omega=(\omega_1,\dots,\omega_m)$, the prover outputs the claim $y := F_T(\omega_1,\dots,\omega_m)$ and proves an opening by iteratively ``peeling off'' variables using univariate polynomial division along each coordinate. Let $r_0(\cdot):=F_T(\cdot)$. For $i\in[m]$, the quotient $q_i$ and remainder $r_i$ polynomials are shown as:
\begin{align}
r_{i-1}&(X_i,\dots,X_m)\nonumber
=\\&q_i(X_i,\dots,X_m)(X_i-\omega_i)+r_i(X_{i+1},\dots,X_m),\label{eq:peel}
\end{align}
so that after $m$ steps the final remainder is the scalar $r_m=y$. The prover commits to each quotient at the trapdoor point by:
\begin{equation}
\pi_{\omega_i} := G^{q_i(\tau_i,\dots,\tau_m)}\in \mathcal{G},
\qquad
\pi_\omega := (\pi_{\omega_1},\dots,\pi_{\omega_m}).\label{eq:proofdef}
\end{equation}
Eq.~\ref{eq:proofdef} yields a proof $\pi_\omega$ without revealing $F_T$ or the rest of $T$.

The opening procedure can be implemented by representing coefficients of $F_T$ in a lexicographic array and running a division for each $i\in[m]$ with overall cost $O(d^m)$ field operations under a uniform per-axis bound $d-1$.

Given $(C_T,\omega,z,\pi_\omega)$, buyer agent $a$ checks whether the acquired asset $T[\omega]=z$ satisfies the pairing equation of the form:
\begin{equation}
e(C_T\cdot G^{-z},\, G) \stackrel{?}{=}
\prod_{j=1}^me\left(\pi_{\omega_j}, G^{(\tau_j-\omega_j)}\right). \label{eq:pairing_check}
\end{equation}

Eq.~\ref{eq:pairing_check} holds when the quotient identity implied by Eq.~\ref{eq:peel} is satisfied at the trapdoor points. Neither the full data nor the polynomial is necessary, as the public SRS enables this check. Once $C$ is fixed, an adversary cannot produce two distinct valid proofs at the same $\omega$. This property is known as positional binding.\\

\begin{figure}[t]
    \centering
    \includegraphics[width=0.9\linewidth]{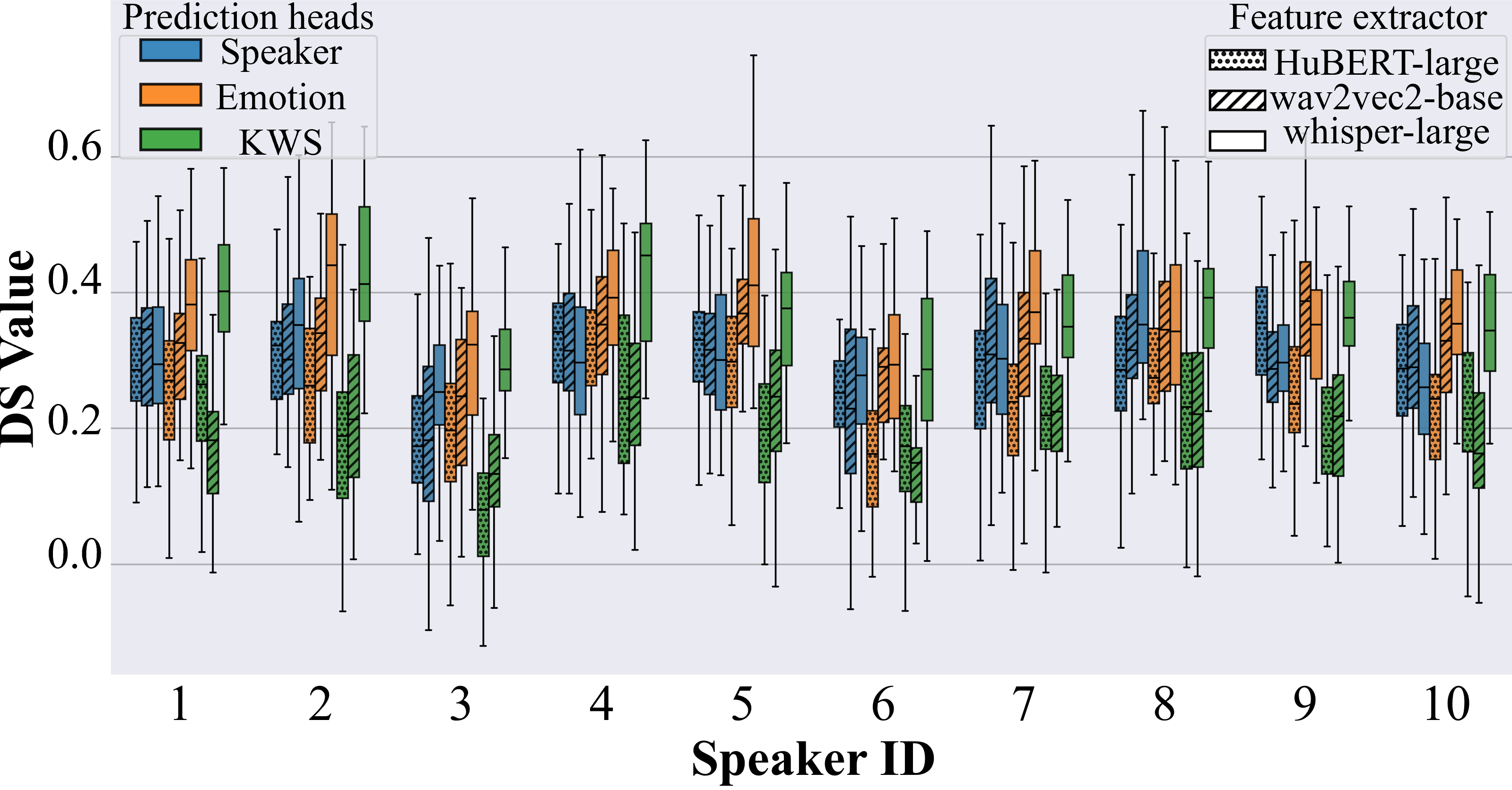}
    \caption{\textbf{Per-Speaker DS for SAM Pricing:} For each speaker, we show DS distributions over \emph{utterance-level} embeddings across three buyer heads (color) and three feature extractors (hatch), computed confidentially in the TEE. This reveals \emph{who} to buy from and \emph{which embedding space} best serves each task.}
    \label{fig:shapleyanalysis}
\end{figure}

\noindent\textbf{\underline{WaveTrees:}} To scale the verification across many assets and support the \emph{hierarchical} organization of a speech database (e.g., seller $\rightarrow$ speaker $\rightarrow$ utterance $\rightarrow$ band $\rightarrow$ embeddings), we use a tensor-native authenticated tree, \emph{WaveTrees} (WT). A WT over tensor $T$ with order $m$ and shape $d=(d_1,\dots,d_m)$ is a $B$-ary tree with $B=\prod_{t=1}^m d_t$, where each internal node $\eta$ at level $\ell$ has children indexed by a multi-index $\textbf{k}_\ell=(k_{\ell,1},\dots,k_{\ell,m})\in \prod_{t=1}^m [d_t]$. The node gathers its children into a tensor $\textbf{Z}_{\eta_\ell}$ whose entries are payloads or child commitments, and stores the commitment $C_{\eta_\ell} := G^{F_{Z_{\eta_
\ell}}(\tau_1,\dots,\tau_m)}$ where the root digest is $C_{\eta_0}.$ For a leaf indexed by $i$ with root-to-leaf path $(\eta_0,\dots,\eta_L)$ for $L$ levels and per-level child multi-indices $\mathbf{k}_\ell$ such that $\eta_{\ell+1}=\eta_\ell[\mathbf{k}_\ell]$, the WT proof sequence is:
\begin{equation}
\pi^{T,i} := (\pi_{\eta_0},\dots,\pi_{\eta_L}).
\qquad
\pi_{\eta_\ell} := (\pi_{k_{\ell,1}},\dots,\pi_{k_{\ell,m}}),
\label{eq:tt_proof}
\end{equation}
The buyer agent verifies at each level from the bottom up until it reaches the root digest $C_{\eta_0}$. At equal arity, this yields fewer levels and cheaper per-node proving than univariate trees.

We use WC to commit speech knowledge tensors (e.g., spectrogram blocks, embedding matrices) and the released valuation tensor $V$, and WT to mirror the natural hierarchy of speech assets. The public bulletin stores only root commitments, while buyers obtain short WC/WT openings only for purchased indices and queried tensors.

\section{Experimental Setup}\label{sec:exp}
\noindent Here, we present our metrics, datasets, encoders, and baselines.\\

\noindent\textbf{\underline{Evaluation Metrics}} involve three market-critical metrics.\\
\textit{(i) Data Shapley:} We compare the buyer utility gains by
Eq.~\ref{eq:shapley_mc}.\\
\textit{(ii) Latency:} We time committing and verifying an asset.\\
\textit{(iii) Proof Size:} We measure the network cost in bytes/proof.\\

\noindent\textbf{\underline{Datasets:}} We use three datasets for diverse representations. \textit{TIMIT} \cite{timit} targets speaker ID with controlled read speech and broad phonetic coverage for clean analysis of $\approx$5h of recordings from 630 speakers, each with 10 utterances. \textit{IEMOCAP}~\cite{iemocap} focuses on emotion recognition in natural dialogs with $\approx$12h of recordings from 5 sessions and 10 speakers, totaling 10{,}039 emotion-labeled utterances. \textit{Speech Commands} (SC)~\cite{speechcommandsv2} targets keyword spotting with 105{,}829 short clips from 2{,}618 speakers over 35 keywords, yielding frequent frame-level openings where size and latency matter most.\\

\noindent\textbf{\underline{Baselines:}} We compare three authenticated substrates for sealing and selective opening of speech marketplace objects.
\textit{Merkle Trees (MTs)} commit to a flat list and open entries via a log-depth hash path (strong integrity, no tensor structure)~\cite{merkle}.
\textit{Univariate Commitment Tree (UCT)} flattens tensors and commits in 1D, increasing depth and losing structure-aware partial openings, motivating a multivariate design~\cite{verkle}.
\textit{WaveTree (WT, ours)} commits assets and valuation tables in native tensor form, yielding fewer levels for the same fanout and cheaper selective openings of structured speech regions.\\

\noindent\textbf{\underline{Encoders:}} To model heterogeneous speech agents, sellers extract embeddings with three foundation models: \textit{Whisper-large}~\cite{whisper}, \textit{HuBERT-large}~\cite{hubertcitation}, \textit{wav2vec~2.0-base}~\cite{wav2vec2citation} with embeddings of sizes 1280, 1024, and 768, respectively. This mix forces the market to value and verify \emph{representation-dependent} assets (e.g., content/prosody-focused vs. speech-to-text-centric features). For each utterance, we compute \textit{window-level} embeddings, average them at the \textit{utterance level}, and then average per speaker to obtain a \textit{speaker-level} voice-biometry.\\

\noindent\textbf{\underline{Prediction Heads:}} Each buyer agent confidentially uploads its lightweight heads $H_k^a(\cdot)$ into the TEE, which defines its private utility $U_a(\cdot)$: \textit{Speaker Identification} (SI), \textit{Emotion Recognition} (ER), and \textit{Keyword Spotting} (KWS) heads. This diversity makes utility head-dependent, ruling out metadata-only pricing and requiring confidential, model-specific valuation.

\section{Results}

Our analysis in Fig.~\ref{fig:shapleyanalysis} shows that \textbf{value is not a property of the dataset alone}. The same speaker's utterances can be highly useful or nearly irrelevant depending on the \textbf{buyer head} and the \textbf{seller's embedding space}. This head- and speaker-specific variation motivates agent-level valuation before any seller-buyer handshake. Table~\ref{tab:timings} then shows that \textbf{verification must also be practical}: WT is the only method that keeps \emph{both} proof bytes and time low enough to support markets proving and verifying hundreds of entries per trade.\\

\noindent\textit{\underline{Which embeddings are worth buying for which task?}}\\
Fig.~\ref{fig:shapleyanalysis} shows speaker value is buyer and encoder dependent. Two trends emerge. \textit{(i)} Representation choice dominates ER and KWS: \textit{Whisper} has the top mean DS (ER: 0.364 vs. 0.324/0.249; KWS: 0.370 vs. 0.197/0.200) and best in both. SI is less encoder-sensitive: means are closer (0.305 vs. 0.291/0.285) and the best encoder varies by speaker. \textit{(ii)} \textit{Whisper} yields no negative-valued utterances (0\% vs. 0.5\%/5.5\%), reducing the risk of paying for useless data.\\

\noindent\textit{\underline{How costly is it to verify in time \emph{and} size at market scale?}}\\
\noindent Table~\ref{tab:timings} answers the practical question for SAMs. We list per-speaker overheads to verify a sealed item across the benchmarks. All methods commit a 32B digest, but the other costs differ sharply across the datasets. MTs pay a $\approx$5.33KB proof per query, which becomes bandwidth-heavy when verifying many embeddings. UCTs compress the proof to $\approx$578B but are \textit{prohibitively slow to commit} ($\approx$9.8s per opening), making large-scale settlement infeasible. We hit the regime SAMs need: \textit{small proofs} ($\approx$138B, \textbf{39.5}$\times$ reduction vs. MT) with \textit{fast commits} ($\approx$0.5s, \textbf{17.5}$\times$ speedup vs. UCT), while keeping verification \textbf{5.1}$\times$ faster than UCT ($\approx$5.6ms). Beyond the means, WT peaks at \textbf{43.2}$\times$ smaller proofs and \textbf{18.1}$\times$ faster commits. Only we deliver \textbf{the smallest proofs} and fairly \textbf{low latency} \underline{simultaneously}.
\begin{table}[t]
\centering
\small
 \caption{Do SAMs validate the assets fast enough to trade at scale?
 \label{tab:timings}}
\setlength{\tabcolsep}{2pt}
\begin{tabular}{c|c|c|c|c}
    & \textbf{Metric} & \textbf{MT} & \textbf{UCT} & \textbf{WT} (Ours) \\
    \midrule
    \multirow{4}{*}{\rotatebox[origin=c]{90}{\textit{TIMIT}}}
        & Commit Size ($\downarrow$)  & 32B  & 32B  & 32B \\
        & Proof Size ($\downarrow$)   & 4280B & 480B & \textbf{128B} \\
        & Commit Time ($\downarrow$)  & 12.2$\pm$2.6ms  & 409$\pm$8.7ms   & \emph{32.6$\pm$2.8ms }\\
        & Verify Time ($\downarrow$)  & $\approx$1ms    & 25.2$\pm$2.2ms  & \emph{5.6$\pm$1.1ms}  \\
    \midrule
    \multirow{4}{*}{\rotatebox[origin=c]{90}{\textit{IEMOCAP}}}
        & Commit Size ($\downarrow$)  & 32B   & 32B  & 32B \\
        & Proof Size ($\downarrow$)   & 6480B & 672B & \textbf{157B} \\
        & Commit Time ($\downarrow$)  & 617$\pm$49ms     & 24544$\pm$324ms  & \textit{1358$\pm$281ms} \\
        & Verify Time ($\downarrow$)  & $\approx$1ms     & 32.3$\pm$1.4ms   & \textit{5.6$\pm$1.6ms}  \\
    \midrule
    \multirow{4}{*}{\rotatebox[origin=c]{90}{\textit{SC}}}
        & Commit Size ($\downarrow$)  & 32B   & 32B  & 32B \\
        & Proof Size ($\downarrow$)   & 5612B & 582B & \textbf{130B} \\
        & Commit Time ($\downarrow$)  & 116$\pm$27.7ms   & 4429$\pm$5454ms  & \textit{293$\pm$48.4ms} \\
        & Verify Time ($\downarrow$)  & $\approx$1ms     & 28.7$\pm$1.0ms   & \textit{5.6$\pm$0.9ms}  \\

\end{tabular}
\end{table}

\section{Conclusion}

In this paper, we introduce WC to make \textit{confidential} SAMs practical by combining \textit{private, model-specific valuation} with \textit{public, selective verification}. As \textit{speech knowledge is tensor-structured}, integrity must be \textit{multivariate}. Tensors are committed once, and queried regions are opened via short proofs, while all else stays hidden. A TEE computes DS per buyer head, and WTs bind them to delivered assets, preventing substitution and plausible forgeries. Our experiments show value strongly depends on head and encoder, supporting agent-level rather than metadata-only pricing. On average, WT achieves $\approx$\textbf{138B} proofs, $\approx$\textbf{0.5s} commitment, and $\approx$\textbf{5.6ms} verification. Proofs are up to \textbf{43.2$\times$ smaller} than those of the fastest prior method, and commitment is up to \textbf{18.1$\times$ faster} than that of the smallest-proof prior method, enabling settlement for thousands of assets. Future work will strengthen provenance against synthetic speech with anti-spoofing and audits, reduce TEE reliance via hybrid cryptography without losing throughput, and support dynamic inventories with updates, revocations, and formal leakage analysis.

\bibliographystyle{IEEEbib}
\bibliography{ref/swarm,ref/mybib}

\end{document}